%% file: iclr2022_conference.tex
\documentclass{article} %

\usepackage{iclr2022_conference,times}

\input{math_commands.tex}

\usepackage{hyperref}
\usepackage{url}
\usepackage{todonotes}
\usepackage{subcaption}

\usepackage[ruled,linesnumbered]{algorithm2e}
\SetKwInput{KwSetup}{Set-up}
\SetKwInput{KwStep}{Step}

\SetCommentSty{mycommfont}

\usepackage{siunitx}
\usepackage{float}
\usepackage{hyperref}

\title{HyperNCA: Growing Developmental \\ Networks  with Neural Cellular Automata}

\author{Elias Najarro,  Shyam Sudhakaran, Claire Glanois, \& Sebastian Risi  \\
Department of Digital Design\\
IT University of Copenhagen\\
Copenhagen, Denmark\\
\texttt{\{enaj,shsu,clgl,sebr\}@itu.dk} \\
}

\iclrfinalcopy
\begin{document}

\maketitle

\begin{abstract}

In contrast to deep reinforcement learning agents, biological neural networks are grown through a self-organized developmental process. Here we propose a new hypernetwork approach to grow artificial neural networks based on neural cellular automata (NCA). Inspired by self-organising systems and information-theoretic approaches to developmental biology, we show that our \emph{HyperNCA} method can grow neural networks capable of solving common reinforcement learning tasks. Finally, we explore how the same approach can be used to build developmental \emph{metamorphosis networks} capable of transforming their weights to solve variations of the initial RL task.

\end{abstract}

\section{Introduction}

Reinforcement learning (RL) agents are either trained with policy gradient methods \citep{arulkumaran2017deep}, or evolved with evolutionary strategies~\citep{Salimans2017Mar} or genetic algorithms \citep{such2017deep,risi2019deep}. In both of these approaches, the neural network weights are directly optimised, meaning that the search space is the weight space. An alternative approach to finding optimal policies are the so-called \textit{indirect encodings} \citep{stanley2019designing,stanley2003taxonomy}. These methods introduce an intermediate step into the training process which decouples the optimisation space from the policy weight space: instead of directly searching for optimal policy weights, we optimise a model whose output is the policy weights. 

A well-known indirect encoding example found in nature is the genotype-phenotype relation between DNA sequences and the proteins they encode. For instance, the information needed to grow a fully-functioning human brain must be contained in the human genome. However, DNA does not encode the final position of every neuron or the presence of every synapse. A simple a back-of-the-envelope calculation shows that the amount of information in the human DNA sequence —approximately 1 GB— is not sufficient to explicitly encode the $10^{15}$ synapses present in the human brain \citep{zador2019critique}. In other words, your brain adjacency matrix is nowhere to be found in your DNA.

Unlike RL agents, animals grow their neural circuitry through a developmental process in which neurons determine their synaptic connections solely based on local interactions. In biological systems, the growth of the nervous system —as any other tissue— is governed by gene regulatory networks: a single set of rules encoded in the cells' DNA which, depending on the cells state, will produce different developmental outcomes.

The fundamental insight, acting as the foundation of this paper, is the connection between the notion of \emph{computational irreducibility} and biological developmental processes. Some dynamical systems —notably chaotic ones, but not only— can have the interesting co-occurrence of two properties: being deterministic and unpredictable simultaneously. In order to determine the state after $n$ steps of such a system, one must evolve the system for $n$ steps according to its dynamical equation; in other words, there does not exist an analytical expression describing the state of the system for any step $n$. Such systems are said to be computationally irreducible \citep{Zwirn2011Nov}. It is conjectured that the developmental process of biological systems is computationally irreducible: the only way to resolve the final configuration of the neural circuitry of the brain is to let unfold the genetic information present in the DNA through an  \textit{algorithmic growth} process \citep{Hiesinger2021Apr}. The power of algorithmic growth resides in not requiring to explicitly codify each of the details of the final configuration of the system. Instead, a small amount of information —which encodes the growth process— is capable of giving rise to a more complex system purely through its self-organised dynamics.

Cellular automata (CA) are computational models able to exhibit computational irreducibility  \citep{Wolfram2002}. They can display complex dynamical patterns emerging exclusively from the local interaction of its elements via simple rules. Recently, neural cellular automata (NCA) —which replace the update rule of the CA by a neural network \citep{wulff1992learning, nichele2017neat, mordvintsev2020growing}— have gained popularity by taking advantage of modern deep learning tools. On the other hand, a recent trend aims to make deep learning systems more robust by combining them with ideas from collective intelligence such as CAs and self-organization  \citep{ha2021collective,risi2021selfassemblingAI}.

Building on insights from algorithmic growth and computational irreducibility, we aim to grow neural networks using neural cellular automata to solve reinforcement learning tasks. We introduce \emph{HyperNCAs}, NCAs acting as Hypernetworks, and show that they are capable of producing policy networks with a significantly larger number of parameters able to solve modern reinforcement learning tasks such as a discrete action environment (e.g.\ Lunar Lander) and a quadrupedal robot locomotion task. Unlike existing Hypernetwork approaches \citep{Ha2016Sep, stanley2009hypercube,carvelli2020evolving}, our proposed indirect-encoding relies on a developmental process guiding the emergence of the final policy network.
Finally, we propose the notion of \textit{metamorphosis networks}: a NCA-based approach to morph neural networks such that the network information is preserved and re-used. We demonstrate how metamorphosis networks can be used to morph the weights of a policy network in order to adapt to differently damaged morphologies. 

The goal of this work is not to reach the ever-elusive state-of-the-art at any given RL tasks, but rather, to introduce a new kind of indirect encoding that puts forward the value of self-organisation and algorithmic growth to generate neural policies for reinforcement learning agents.

\section{Related work}

\textbf{Indirect encodings.} Indirect encodings are inspired by the biological  process of mapping a compact genotype to a larger phenotype and have been primarily studied in the context of neuroevolution \citep{floreanoDuerrMattiussi2008} (i.e.\ evolving artificial neural networks) but more recently were also optimized through gradient-descent based approaches \citep{Ha2016Sep}. In indirect encodings,   the description of the solution is compressed, allowing information to be reused and the final solution to contain more components than the description itself. Even before the success of deep RL, these methods enabled solving challenging car navigation tasks from pixels alone \citep{koutnik2013evolving}.

A highly influential indirect encoding is  HyperNEAT~\citep{stanley2009hypercube}. HyperNEAT employs an indirect encoding called compositional pattern producing networks (CPPNs) that \emph{abstracts away the process of growth} and instead describes the connectivity of a neural network through a function of its geometry. An approach building on this idea, called Hypernetworks \citep{Ha2016Sep}, has more recently shown that networks generating the weights of another network can also be trained end-to-end through gradient descent. Hypernetworks have been shown to be able to generate competitive CNNs \citep{zhmoginov2022hypertransformer} and RNNs in a variety of tasks while using a smaller set of trainable parameters.  Both HyperNEAT and Hypernetworks are indirect encodings that have been applied to reinforcement learning problems \citep{stanley2009hypercube,carvelli2020evolving}, including robot locomotion \citep{risi2013confronting,clune2009evolving}. However, they do not rely on the process of development over time, which can increase the evolvability of artificial agents \citep{kriegman2017minimal,bongard2011morphological} and is an important ingredient of biological systems \citep{Hiesinger2021Apr}.

\textbf{Neural cellular automata (NCA)}. Cellular automata are a class of computational models whose outcomes emerge from the local interactions of simple rules. Introduced by  \cite{neumann1966theory} as a part of his quest to build a self-replicating machine or \textit{universal constructor}, a CA consist of a lattice of computing cells which iteratively update their states based exclusively on their own state and the states of their local neighbours. On a classical CA, each cell's state is represented by an integer and adjacent cells are considered neighbours. Critically, the update rule of a cellular automaton is identical for all the cells.

\citet{wulff1992learning} showed that a recurrent single-layer neural networks could be trained with the delta-rule algorithm to approximate the dynamical patterns generated by chaotic one-dimensional CA. While Wulff and Hertz work was the first one to combine neural networks with cellular automata, they restricted their focus to a top-down study of the CA's dynamics with a neural network rather than replacing the CA rule by a neural neural network and studying their bottom-up emerging properties. \citet{Tavares2006NeuroCellularAC} proposed a Neuro-CA model which replaces the symbolic if-then rules of classical CA by a neural network and represents the states by real numbers rather than integers. However, similarly to \citet{wulff1992learning}, the scope of the work was limited to showing that a small feedfoward NCA with only 8 neurons trained through backprop could reproduce one-dimensional CA patterns such as the Turing complete \textit{Rule 110}.
\citet{Gilpin2018Sep} formulated NCA as a convolutional network architecture and showed that they could model complex dynamical CA patterns such as Conway's Game of Life. Critically, by formulating a NCA as a convolutional neural network, Gilpin enabled NCA to be easily trainable exploiting  modern deep learning techniques.

\cite{mordvintsev2020growing} demonstrated how NCAs can work as robust models of morphogenesis, i.e.\ the process by which single stem cells grow into fully-formed organisms. In their work, the goal of the NCA is to learn to grow 2D images. Each cell state is represented as a 16-dimensional vector with the first 3 channels representing the visible colours, one channel representing the living state of the cell, and the remaining hidden channels representing undefined quantities that the NCA can exploit to encode information.
Their neural rule encompasses a convolutional architecture with Sobel filters as kernels which compute spatial gradients of the cells states, before feeding the resulting activations to feed-forward layers whose output determine the new state of each cell. Our work builds upon this architecture, and similarly to previous work on 3D NCAs by  \cite{Sudhakaran2021Mar},  the convolutional layers are trained along with the rest of the network parameters.

Because of their emergent nature, CAs and NCAs are capable of generating complex dynamical patterns with few model parameters as shown by recent work \citep{Kaiser2015Nov, mordvintsev2020growing, Ruiz2020Jun, Sudhakaran2021Mar}.

\section{Hyper Neural Cellular Automata}

Our \emph{Hypernetwork Neural Cellular Automata}  (HyperNCA) is visually summarised in Figure \ref{fig:main_figure}:

\begin{figure}[H]
         \centering
         \includegraphics[width=1.0\textwidth]{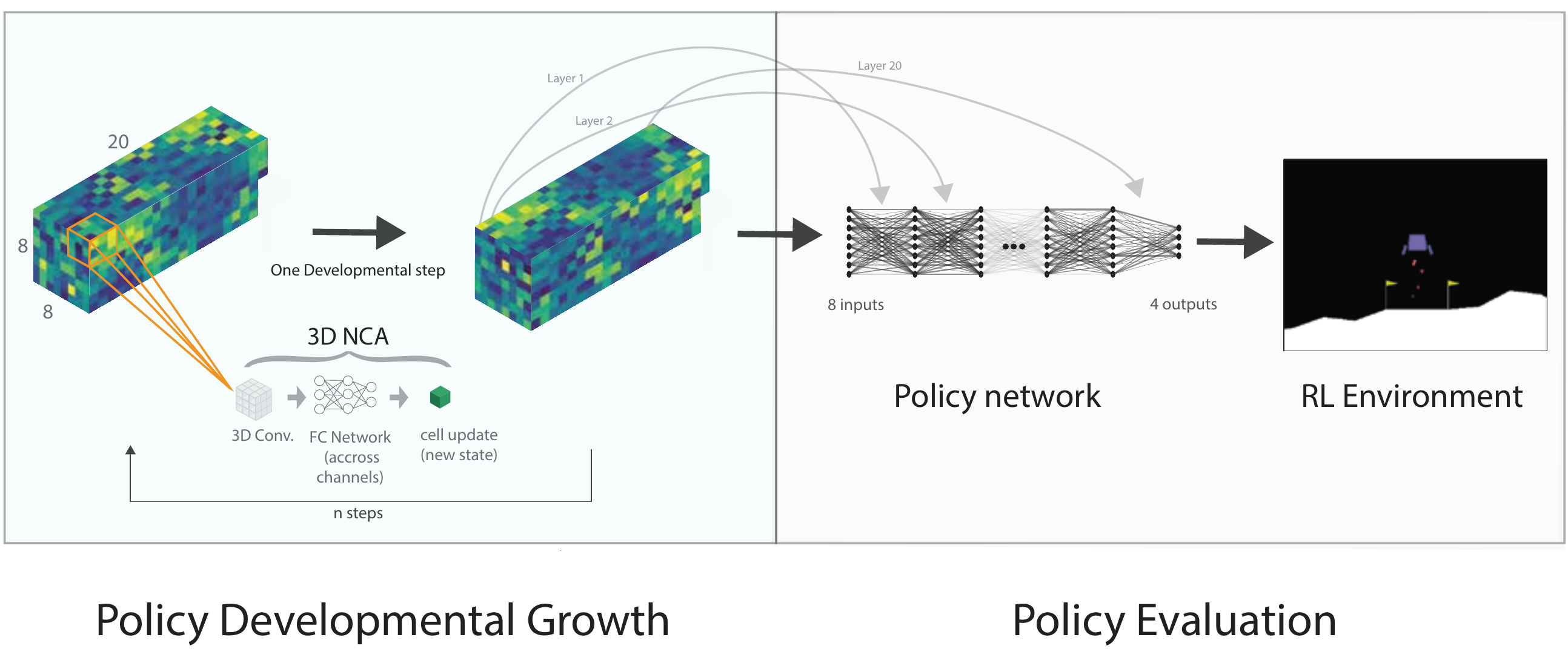}
        \caption{\textbf{Hyper Neural Cellular Automata (HyperNCA).} During the \textbf{Developmental Growth phase} \textit{(left)}, an initial random seed is updated for a finite number of steps using a 3D Neural Cellular Automata (NCA). The NCA —and the seed— can have a single or several channels where information is stored, here a single channel is represented.
        \textbf{Policy evaluation phase} \textit{(right)}: Once development is concluded, the first channel of the grown pattern is mapped to the weight matrix of a policy network which is evaluated on a reinforcement learning task —in this example, safely landing the spacecraft on the ground.
        The size of the seed is chosen such that it matches the size of the RL agent observation space while non-matching values of the last layer are simply zeroed-out. The number of hidden layers of the policy can be arbitrarily chosen using a bigger seed, in the shown example the seed/policy has 20 layers.}
        \label{fig:main_figure}
\end{figure}

The \emph{HyperNCA} approach consists of a neural cellular automata (NCA) which operates on a high-dimensional substrate whose values are interpreted as the weights of another network. We use the NCA as a mathematical model analogous to development in biological organisms.
We do so by training a NCA to guide the developmental process of a policy network that controls a reinforcement learning agent. Similarly to animal development, the growth process of the policy network is a self-organising process whose final configuration emerges solely from local interactions.

Based on previous work \citep{mordvintsev2020growing, Sudhakaran2021Mar}, our  NCA uses a convolutional architecture which operates on a $3$-dimensional substrate where each of the three dimensions has $C$ channels, resulting in a substrate of dimension $3C$. The NCA consists of $3$D convolutions followed by a dense layer. The convolutions have kernel size $3$, hence preserving the locality of the cellular automata transition rule.

In order to generate a policy network we proceed as follows: \textbf{1.} We randomly initialise a substrate sampling from a random uniform distribution. \textbf{2.} The NCA is applied to the substrate for a finite number of steps. \textbf{3.} The values of one of the channels of the substrate are interpreted as the weights of a policy network $\mathcal{P}$. \textbf{4.} The resulting policy $\mathcal{P}$ is evaluated in a reinforcement learning task $\mathcal{T}$. \textbf{5.} The resulting fitness of $\mathcal{P}$ in task $\mathcal{T}$ guides the evolution of the NCA  weights.
The substrate has shape $L \times C \times W \times W$, where $L$ is the number of layers of the policy network, $C$ is the number of channels of the NCA convolutional layers, and $W$ is the size of the observation space of the task (or the action space size if it is bigger than the observation space). This generic approach enables us to grow networks of any size —in depth and width— by simply adjusting the shape of the substrate.
The algorithmic description of the approach is provided in Algorithm \ref{algohypernca}.

\begin{algorithm}[H]
\SetAlgoLined
\vspace{1mm}
\KwIn{Reinforcement learning task $\mathcal{T}$, NCA model, number of developmental steps $\delta$, \\ training hyper-parameters $\Omega$, fitness function $\mathcal{F}$.}
\KwOut{Optimal agent policy weights $\mathcal{W}$.} 
\vspace{1mm}
Sample random substrate seed $\mathcal{S}$ from an uniform distribution $\mathcal{U}(-1,1)$ ; \\
\While{Fitness $\mathcal{F} <$ target fitness}{ 
    \While{generation $<$ generations limit}
    {
    Generate a population of NCA using CMA-ES sampler; \\
        \For(\tcp*[f]{Parallelized across CPU cores}){NCA \textbf{in} NCA population} 
        {  
            Let NCA update the initial seed for $\delta$ steps; \\
            
            The grown values of the first channel are interpreted as the weights of the agent policy $\mathcal{W}$;
            
            The grown policy is evaluated on the RL task $\mathcal{T}$; \\
        }
        
    \If{mean population fitness $\mathcal{F} <$ early-stopping threshold $\in \Omega$ }{
        break\; 
    }
    
    Use population fitness values to update  CMA-ES sampler's distribution mean and covariance matrix;\\
    
    }
    Save solution and mean solution with highest fitness;
} 

 \caption{HyperNCA: Growing neural networks with Neural Cellular Automata \& CMAES}
 \label{algohypernca}
\end{algorithm}

\section{Experiments and results}

\subsection{Optimisation details}

We use CMA-ES —Covariance Matrix Adaptation Evolution Strategy— \citep{Hansen1996May}, a black-box population-based optimisation algorithm to train the HyperNCA. Evolutionary strategies (ES) algorithms have been shown to be capable of finding performing solutions for reinforcement learning tasks, both through directly optimising the policy weights \citep{Salimans2017Mar} and with encodings of the policy in the form of local learning rules \citep{Najarro2020Jul}. Black-box methods such as CMA-ES have the advantage of not requiring to compute gradients and being easily parallelizable. 

Experiments have been run on a single machine with a \textit{AMD Ryzen Threadripper 3990X} CPU with $64$ cores. We choose a population size of $64$, such that we can run one generation in parallel, —each core running an evaluation of the environment—, and adopt an initial variance of $0.1$. Finally, we employ an early stopping method which stops and resets training runs which show poor performance after a few hundred generations.

All the code to train the HyperNCA model in any gym environment as well as the the weights of the reported models can be found at  \url{https://github.com/enajx/hyperNCA}

\subsection{Hypernetwork baseline}
As a baseline, we use a Linear Hypernetwork, similar to the one  in \citet{Ha2016Sep}. The smallest Hypernetwork is composed of a set of linear embeddings and a single linear weight generating matrix. The number of trainable parameters can be calculated by ($N*E_{dim} + E_{dim}*M$, where $N$ equals the number of embeddings and $M = num\_target\_params / N$. Typically, each embedding in the Hypernetwork corresponds to a layer in the target network. However, because the target networks are small ($288$ and $1792$ parameters), we opt to include many embeddings and concatenate the weight generator outputs to create the final parameter vector, to ensure a smaller Hypernetwork.

\subsection{Results}

The HyperNCA method is able of generating policy networks capable of solving tasks with both continuous and discrete action spaces, namely: a simulated 3D quadruped whose goal is to learn to walk as far as possible along one direction and Atari's \emph{Lunar Lander}, a discrete task where the goal is to smoothly land on a procedurally generated terrain.

\textbf{Continuous 3D locomotion task.} This task consists of a simulated robot  using the Bullet physics engine \citep{coumans2021}; the quadruped has $13$ rigid links, including four legs and a torso, along with $8$ actuated joints \citep{Duan2016Apr}. It is modeled after the ant robot in the MuJoCo simulator and constitutes a common benchmark in RL \citep{Finn2017Jul}. The robot has an input size of $28$, comprising the positional and velocity information of the agent and an action space of $8$ dimensions, controlling the motion of each of the $8$ joints. The fitness of the quadruped agent depends on the distance travelled during a period of $1,000$ time-steps along a fixed axis. In order to make our experiment reproducible —and conveniently, easier for the agent— , we made the environment deterministic by fixing the seed which instantiates the robot on the same initial position at each episode.

We follow the method described in Algorithm \ref{algohypernca} to grow a feed forward policy network consisting of three layers with $[28, 28, 8]$ neurons respectively, no bias and  hyperbolic tangent as activation function. This policy network has $1,792$ weight parameters. By contrast, the HyperNCA used to grow the policy has $314$ trainable parameters, roughly $5.7$ times less parameters than the resulting policy network. The grown network yields a reward of $1,192$ meters walked. We have not systematically performed a hyperparameter search of the CMA-ES parameters nor the HyperNCA architecture, therefore it is likely that higher rewards could be reached. However, since these tasks can easily be solved with modern RL methods, fine-tuning hyperparameters to obtain the highest reward is not within the focus of this paper.

In comparison, the small baseline Hypernetwork has $704$ parameters and achieves a lower score of $1078$. We also evaluated a large Hypernetwork with $5,280$ parameters using ES \citep{Salimans2017Mar}. Interestingly, here the network achieves a fitness $> 1,400$, which demonstrates that the Hypernetwork can in fact perform well with a large enough number of parameters but is challenged to compress the weights to the same extent than the NCA. Future work will have to investigate these results closer to draw definite conclusions.

\begin{figure}[H]
         \centering
         \includegraphics[width=1.0\textwidth]{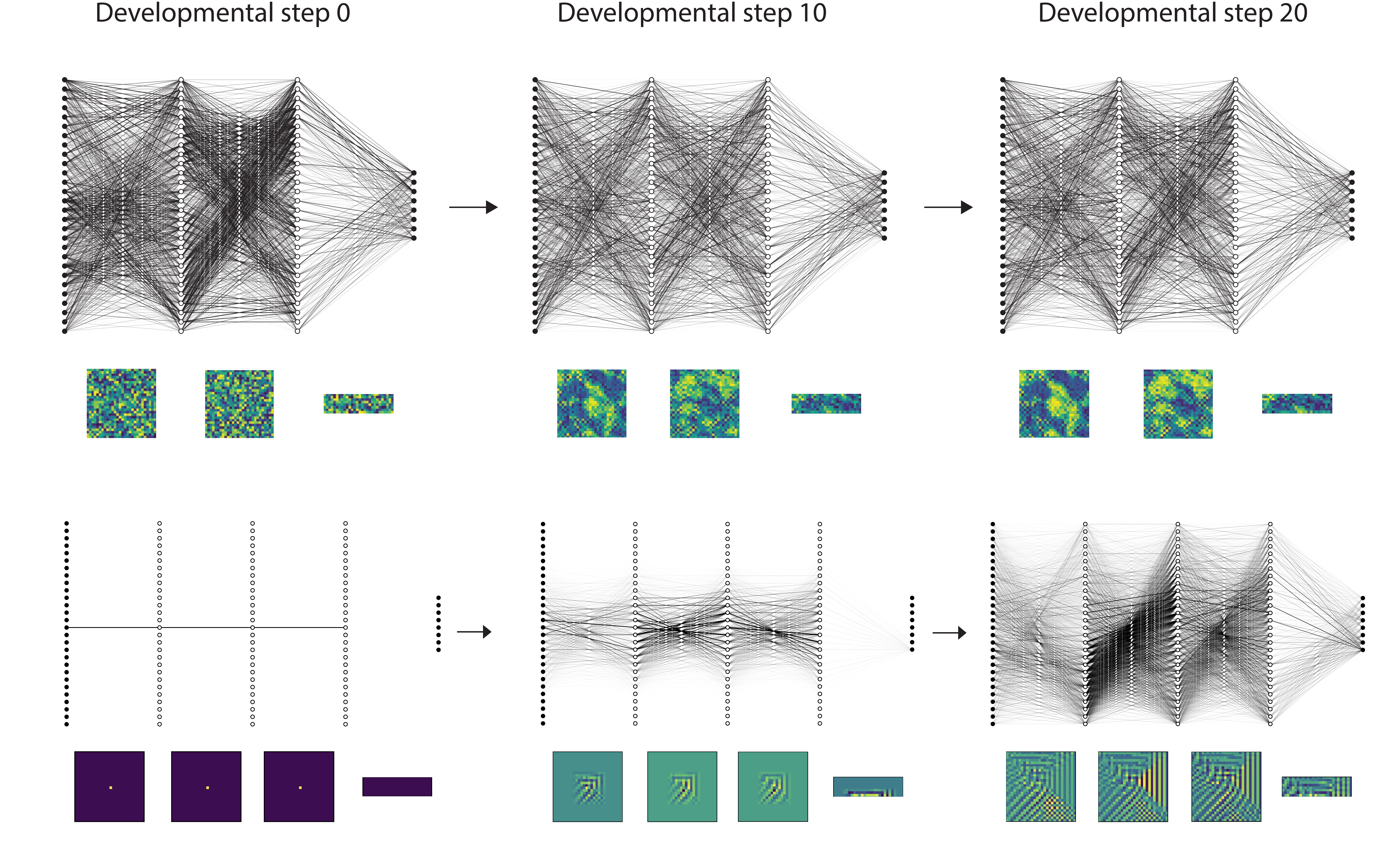}
        \caption{The developmental process of two quadruped policy networks at three developmental steps. \textit{Above:} a network grown from an initial random substrate, after $10$ and after $20$ developmental steps.  \textit{Below}: a network grown from a substrate initialised with a single value at its center. Below each layer, the connections are visualised as weight matrices. 
        }
        \label{fig:devsteps}
\end{figure}

Figure~\ref{fig:devsteps} shows an example of the developmental steps for a policy network grown by an evolved HyperNCA. The weight pattern quickly self-organizes into a particular global structure  after which it appears to be mostly fine-tuned for the remaining developmental steps.

\textbf{Growing deeper policy networks.} In order to demonstrate the flexibility and potential of our approach, we generate a deeper policy consisting of $30$ hidden layers and $22,960$ parameters  by an evolved HyperNCA with $548$ trainable parameters, that is, a phenotype $\sim$40 times bigger than its genotype. The resulting policy reaches a fitness reward of $1,075$. While such a big policy is not necessary to solve the quadruped task, this result demonstrates that the developmental compression exhibited by the HyperNCA can encode high-dimensional phenotypes. Crucially, this enables the optimisation algorithm to operate on a much smaller parameter space, allowing us to rely upon optimisation techniques —like covariance-based  approaches— that would struggle with memory issues in higher dimensional spaces.

\textbf{Discrete RL task.} \textit{Lunar lander} is an single-player Atari game where the player's goal is to smoothly land the lander on a procedurally generated terrain.  There are four discrete actions available: do nothing, fire engine towards the left, fire engine vertically and fire engine towards the right. The environment state is defined by $8$ quantities: the $2$D position coordinates of the lander, its linear velocities, its angle, its angular velocity, and two booleans representing whether each leg is in contact with the ground or not \citep{Brockman2016Jun}. The reward for moving from the top of the screen to the landing pad and coming to rest is about $100$-$140$ points.
If the lander moves away from the landing pad, it loses reward.
If the lander crashes, it receives an additional $-100$ points. If it comes to rest, it receives an additional $+100$ points. Each leg with ground contact is $+10$ points.
Firing the main engine is $-0.3$ points each frame. Firing the side engines $-0.03$ points each frame. The task is considered solved if the player attains $200+$ reward over $100$ evaluations.

To solve this task, we grow a five-layer feedforward policy with $[8, 8, 8, 8, 3]$ nodes per layer respectively, no bias and hyperbolic tangent as activation function.
The resulting policy has $288$ parameters and is generated by a HyperNCA with $226$ trainable parameters, i.e.\ a HyperNCA genotype with roughly $24\%$ less parameters than the resulting policy phenotype.  The policy is grown for $20$ developmental steps and yields a reward of $252 \pm 63$ evaluated over one hundred runs, therefore solving the task according to the pre-establish criterion of $200$ points. Similarly to the previous task, the hyperparameters have not been optimised. For comparison, in this domain the baseline Hypernetwork has $252$ trainable parameters and achieves $264 \pm 28$ fitness over $100$ trials.

\subsection{Metamorphosis neural networks}

Metamorphosis is a developmental phenomenon by which an organism experiences substantial changes in its morphology during post-embryonic life —a biological process first appeared in pterygote insects $400$ million years ago \citep{metamorph}. We draw inspiration from this biological process to build \emph{metamorphosis networks}: an extension of our HyperNCA approach to morph policy networks in order to make the agent capable of solving different tasks. We demonstrate our approach with the quadruped RL task described in the previous section.

First, following the standard HyperNCA approach, a NCA is trained to grow a policy network for the  quadruped locomotion task with standard morphology. Subsequently, the morphology is modified, and the same NCA is let to operate on the grown policy for a finite number of steps in order to adapt its weights to the new morphology. This new \textit{metamorphosed} policy is then evaluated on the task with the altered quadruped body. Finally, we repeat the same metamorphosis process for a third morphology. The metamorphosis of the policy happens \textit{offline}, unlike plastic networks, all the information used to modify the policy weights to the new morphology is contained in the genome (the NCA) and hence no lifelong learning from the environment takes place.

The three quadruped variations consist of: the standard Ant quadruped \textit{M1} and two damaged morphologies \textit{M2} and \textit{M3} with partially mutilated legs. The trained NCA develops a functional policy out of the initial seed in $10$ developmental steps which solves the first morphology \textit{M1}. We then let the same NCA further modify the previously grown weights for $20$ additional developmental steps, and evaluate the resulting network on the second morphology \textit{M2}. The same procedure is applied for the third morphology, where the weights develop for another $20$ steps  to solve morphology \textit{M3}. The resulting policy networks yield $1,329$, $1,343$ and $1,266$ rewards (distance walked) on each morphology respectively. The evolved NCA has $1,480$ trainable parameters, while each of the policy it develops has $1,792$ parameters. Notice that it is a single NCA model that give rise to the three final policy networks. 

\begin{figure}[h!]
         \centering
         \includegraphics[width=0.8\textwidth]{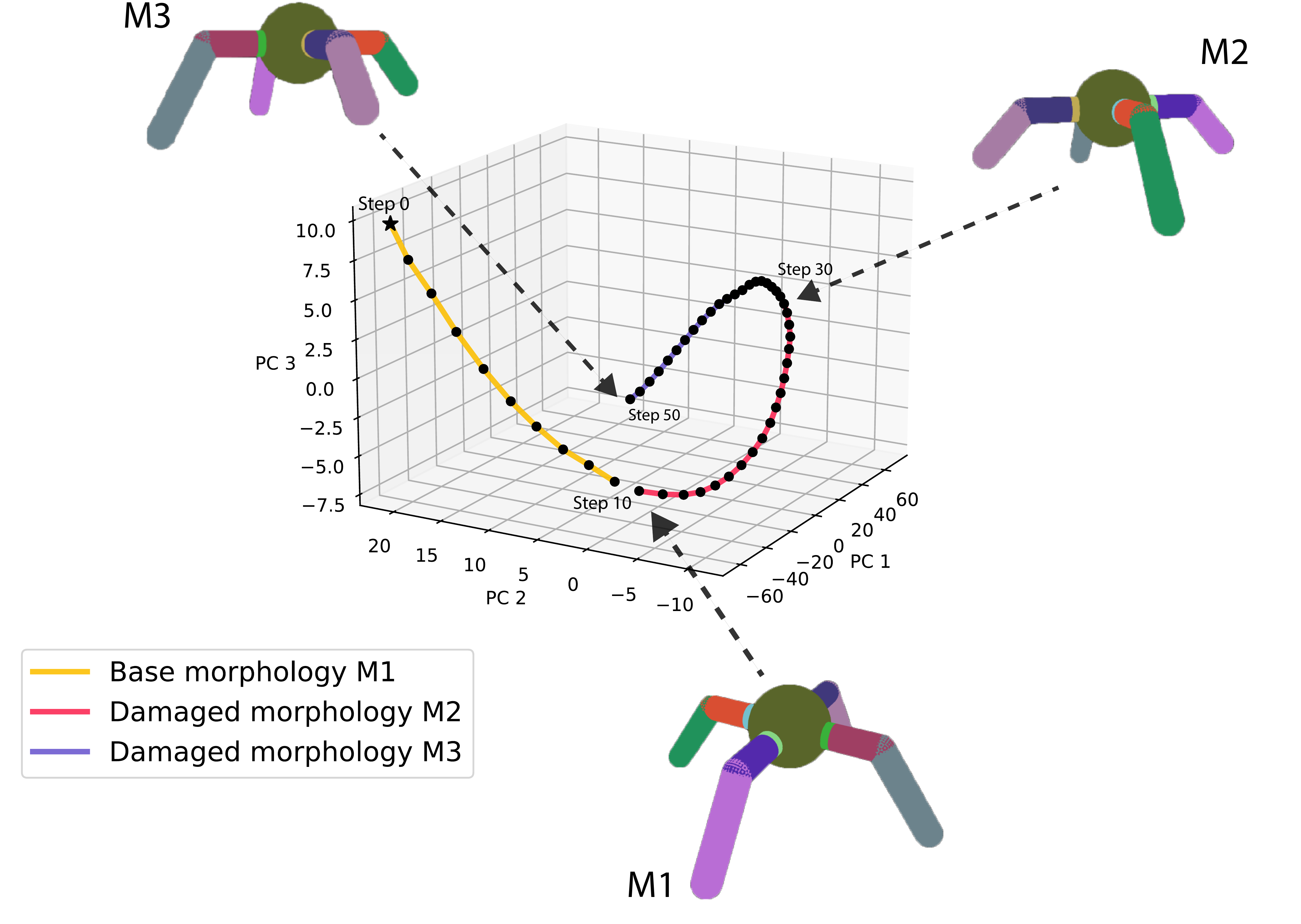}
        \caption{Low-dimensional representation of the policy weights at each developmental step. Each morphology (\textit{M1}, \textit{M2}, and \textit{M3}) points to the location in weight space used by the agent during evaluation. M1 is the standard Mujoco Ant morphology, M2 has one front leg and back leg mutilated and M3 has the two front legs mutilated. The dimensionality reduction is computed with PCA. The initial seed is represented with a star symbol $\star$.
        }
        \label{fig:metamorphosis}
\end{figure}

In order to demonstrate that the policy weights are indeed morphing (i.e.\ the model is not using the same weights to solve all three morphologies), we show in Fig.~\ref{fig:metamorphosis} the relative changes in the weights between each policy as the weight trajectories represented in 3-dimensional PCA space. 

Additionally, we cross-evaluate each of the policies on the mismatching morphologies. Table~\ref{tablecross} shows that the NCA-guided metamorphosis process has indeed a positive impact on the policies' performance on each morphology reward.

\begin{table}[h]
\begin{center}
\begin{tabular}{||c c c c||} 
 \hline
  & M1 & M2 & M3 \\ [0.5ex] 
 \hline\hline
 Policy M1 &  \textbf{1,329} & 827 & 747 \\ 
 \hline
 Policy M2 & 21 & \textbf{1,343} & 916 \\
 \hline
 Policy M3 & 13 & 1,140 & \textbf{1,266} \\
 \hline
\end{tabular}
\end{center}
\caption{Rewards obtained from cross-evaluation of the three policies grown by the same HyperNCA at three developmental steps on each of the three morphologies M1, M2 and M3. The reward value represent the distance walked by the quadruped from its initial position. The values in the diagonal (shown in bold) correspond to the rewards the policies were grown to perform well on, i.e.\ policy M1 was evaluated on morphology M1 at developmental step 10, policy M2 was evaluated on morphology M2 at developmental step 30 and policy M3 was evaluated on M3 at step 50.
That is, the diagonal values are the rewards obtained by each quadruped morphology at their corresponding developmental steps.
The values outside of the diagonal are the rewards obtained when evaluating the policies at the developmental steps which do not match the morphologies at that step; this decreases the obtained reward demonstrating that the policy weights have indeed  morphed.}
\label{tablecross}
\end{table}

\section{Discussion and Future work}

Our HyperNCA approach is able to solve the tested RL tasks, reaching a comparable performance to other indirect encoding methods. Since the model has to undergo a developmental process, it seems currently harder to optimise, although more experiments are necessary to confirm this hypothesis. To optimize these highly dynamical and self-organizing systems, more open-ended search methods such as quality diversity (QD) \citep{pugh2016quality} or intrinsically-motivated learning approaches \citep{baranes2013active} could be of particular interest. These methods might be more suited to navigate the likely highly deceptive fitness landscapes of self-organizing systems.

The goal of this paper is not necessarily to reach SotA in one RL benchmark, but rather to propose a novel developmental framework to grow neural networks through local self-organising processes. Contrasting to the previous literature on Hypernetworks and indirect encodings, our encoding follows a developmental process, inspired by biological growth and self-organisation. While currently not outperforming direct encoding approaches, research in neuroscience suggests that this type of ``genomic bottleneck'' \citep{zador2019critique} has an important regularizing constraint, allowing animals to generalize better to new situations. Recent results suggest that artificial agents can also benefit from such a compression for better generalization \citep{pedersen2021evolving} and the HyperNCA approach presented in this paper could be a promising method to study this question in more detail. Furthermore, it could serve as a model to characterize developmental processes in biological organisms.  

Future work includes extending the approach to more complex tasks, which might benefit from the HyperNCA's ability to grow deep policy networks more clearly. Additionally, it should be explored how different substrate representations —including higher dimensionality— affect the growth process. We are particularly excited to further investigate  the metamorphosis capabilities of such developmental encodings. For example, the changes in an organism's form in biological metamorphosis are often striking (e.g.\ a caterpillar turning into a butterfly). Could a HyperNCA facilitate similar radical changes in the morphologies and nervous systems of artificial robots?

\section*{Acknowledgments}
This project was supported by a Sapere Aude: DFF-Starting Grant (9063-00046B) and GoodAI Research Award.

\newpage

\bibliography{iclr2022_conference}
\bibliographystyle{iclr2022_conference}

\end{document}

%% file: math_commands.tex
\usepackage{amsmath,amsfonts,bm}

\def\eqref#1{equation~\ref{#1}}

\def\1{\bm{1}}

\DeclareMathAlphabet{\mathsfit}{\encodingdefault}{\sfdefault}{m}{sl}
\SetMathAlphabet{\mathsfit}{bold}{\encodingdefault}{\sfdefault}{bx}{n}

%% file: iclr2022_conference.bbl
\begin{thebibliography}{41}
\providecommand{\natexlab}[1]{#1}
\providecommand{\url}[1]{\texttt{#1}}
\expandafter\ifx\csname urlstyle\endcsname\relax
  \providecommand{\doi}[1]{doi: #1}\else
  \providecommand{\doi}{doi: \begingroup \urlstyle{rm}\Url}\fi

\bibitem[Arulkumaran et~al.(2017)Arulkumaran, Deisenroth, Brundage, and
  Bharath]{arulkumaran2017deep}
Kai Arulkumaran, Marc~Peter Deisenroth, Miles Brundage, and Anil~Anthony
  Bharath.
\newblock Deep reinforcement learning: A brief survey.
\newblock \emph{IEEE Signal Processing Magazine}, 34\penalty0 (6):\penalty0
  26--38, 2017.

\bibitem[Baranes \& Oudeyer(2013)Baranes and Oudeyer]{baranes2013active}
Adrien Baranes and Pierre-Yves Oudeyer.
\newblock Active learning of inverse models with intrinsically motivated goal
  exploration in robots.
\newblock \emph{Robotics and Autonomous Systems}, 61\penalty0 (1):\penalty0
  49--73, 2013.

\bibitem[Belles(2020)]{metamorph}
Xavier Belles.
\newblock \emph{Insect Metamorphosis: From Natural History to Regulation of
  Development and Evolution}.
\newblock Academic Pr, 1 edition, 2020.
\newblock ISBN 0128130202; 9780128130209.

\bibitem[Bongard(2011)]{bongard2011morphological}
Josh Bongard.
\newblock Morphological change in machines accelerates the evolution of robust
  behavior.
\newblock \emph{Proceedings of the National Academy of Sciences}, 108\penalty0
  (4):\penalty0 1234--1239, 2011.

\bibitem[Brockman et~al.(2016)Brockman, Cheung, Pettersson, Schneider,
  Schulman, Tang, and Zaremba]{Brockman2016Jun}
Greg Brockman, Vicki Cheung, Ludwig Pettersson, Jonas Schneider, John Schulman,
  Jie Tang, and Wojciech Zaremba.
\newblock {OpenAI Gym}.
\newblock \emph{arXiv}, Jun 2016.
\newblock URL \url{https://arxiv.org/abs/1606.01540v1}.

\bibitem[Carvelli et~al.(2020)Carvelli, Grbic, and Risi]{carvelli2020evolving}
Christian Carvelli, Djordje Grbic, and Sebastian Risi.
\newblock Evolving hypernetworks for game-playing agents.
\newblock In \emph{Proceedings of the 2020 Genetic and Evolutionary Computation
  Conference Companion}, pp.\  71--72, 2020.

\bibitem[Clune et~al.(2009)Clune, Beckmann, Ofria, and
  Pennock]{clune2009evolving}
Jeff Clune, Benjamin~E Beckmann, Charles Ofria, and Robert~T Pennock.
\newblock Evolving coordinated quadruped gaits with the hyperneat generative
  encoding.
\newblock In \emph{2009 iEEE congress on evolutionary computation}, pp.\
  2764--2771. IEEE, 2009.

\bibitem[Coumans \& Bai(2016)Coumans and Bai]{coumans2021}
Erwin Coumans and Yunfei Bai.
\newblock Pybullet, a python module for physics simulation for games, robotics
  and machine learning.
\newblock \url{http://pybullet.org}, 2016.

\bibitem[Duan et~al.(2016)Duan, Chen, Houthooft, Schulman, and
  Abbeel]{Duan2016Apr}
Yan Duan, Xi~Chen, Rein Houthooft, John Schulman, and Pieter Abbeel.
\newblock {Benchmarking Deep Reinforcement Learning for Continuous Control}.
\newblock \emph{arXiv}, Apr 2016.
\newblock URL \url{https://arxiv.org/abs/1604.06778v3}.

\bibitem[Finn et~al.(2017)Finn, Abbeel, and Levine]{Finn2017Jul}
Chelsea Finn, Pieter Abbeel, and Sergey Levine.
\newblock {Model-Agnostic Meta-Learning for Fast Adaptation of Deep Networks}.
\newblock In \emph{{International Conference on Machine Learning}}, pp.\
  1126--1135. PMLR, Jul 2017.
\newblock URL \url{https://proceedings.mlr.press/v70/finn17a.html}.

\bibitem[Floreano et~al.(2008)Floreano, D\"urr, and
  Mattiussi]{floreanoDuerrMattiussi2008}
Dario Floreano, Peter D\"urr, and Claudio Mattiussi.
\newblock Neuroevolution: from architectures to learning.
\newblock \emph{Evolutionary Intelligence}, 1\penalty0 (1):\penalty0 47--62,
  2008.

\bibitem[Gilpin(2018)]{Gilpin2018Sep}
William Gilpin.
\newblock {Cellular automata as convolutional neural networks}.
\newblock \emph{arXiv}, Sep 2018.
\newblock URL \url{https://arxiv.org/abs/1809.02942}.

\bibitem[Ha \& Tang(2021)Ha and Tang]{ha2021collective}
David Ha and Yujin Tang.
\newblock Collective intelligence for deep learning: A survey of recent
  developments.
\newblock \emph{arXiv preprint arXiv:2111.14377}, 2021.

\bibitem[Ha et~al.(2016)Ha, Dai, and Le]{Ha2016Sep}
David Ha, Andrew Dai, and Quoc~V. Le.
\newblock {HyperNetworks}.
\newblock \emph{arXiv}, Sep 2016.
\newblock URL \url{https://arxiv.org/abs/1609.09106v4}.

\bibitem[Hansen \& Ostermeier(1996)Hansen and Ostermeier]{Hansen1996May}
N.~Hansen and A.~Ostermeier.
\newblock {Adapting arbitrary normal mutation distributions in evolution
  strategies: the covariance matrix adaptation}.
\newblock In \emph{{Proceedings of IEEE International Conference on
  Evolutionary Computation}}, pp.\  312--317. IEEE, May 1996.
\newblock ISBN 978-0-7803-2902.
\newblock URL \url{https://ieeexplore.ieee.org/document/542381}.

\bibitem[Hiesinger(2021)]{Hiesinger2021Apr}
Peter~Robin Hiesinger.
\newblock \emph{{The Self-Assembling Brain}}.
\newblock Princeton University Press, Princeton, NJ, USA, Apr 2021.
\newblock ISBN 978-0-69118122-6.
\newblock URL
  \url{https://press.princeton.edu/books/hardcover/9780691181226/the-self-assembling-brain}.

\bibitem[Kaiser \& Sutskever(2015)Kaiser and Sutskever]{Kaiser2015Nov}
{\L}ukasz Kaiser and Ilya Sutskever.
\newblock {Neural GPUs Learn Algorithms}.
\newblock \emph{arXiv}, Nov 2015.
\newblock URL \url{https://arxiv.org/abs/1511.08228v3}.

\bibitem[Koutn{\'\i}k et~al.(2013)Koutn{\'\i}k, Cuccu, Schmidhuber, and
  Gomez]{koutnik2013evolving}
Jan Koutn{\'\i}k, Giuseppe Cuccu, J{\"u}rgen Schmidhuber, and Faustino Gomez.
\newblock Evolving large-scale neural networks for vision-based reinforcement
  learning.
\newblock In \emph{Proceedings of the 15th annual conference on Genetic and
  evolutionary computation}, pp.\  1061--1068, 2013.

\bibitem[Kriegman et~al.(2017)Kriegman, Cheney, Corucci, and
  Bongard]{kriegman2017minimal}
Sam Kriegman, Nick Cheney, Francesco Corucci, and Josh~C Bongard.
\newblock A minimal developmental model can increase evolvability in soft
  robots.
\newblock In \emph{Proceedings of the Genetic and Evolutionary Computation
  Conference}, pp.\  131--138, 2017.

\bibitem[Mordvintsev et~al.(2020)Mordvintsev, Randazzo, Niklasson, and
  Levin]{mordvintsev2020growing}
Alexander Mordvintsev, Ettore Randazzo, Eyvind Niklasson, and Michael Levin.
\newblock Growing neural cellular automata.
\newblock \emph{Distill}, 2020.
\newblock \doi{10.23915/distill.00023}.
\newblock URL \url{https://distill.pub/2020/growing-ca}.

\bibitem[Najarro \& Risi(2020)Najarro and Risi]{Najarro2020Jul}
Elias Najarro and Sebastian Risi.
\newblock {Meta-Learning through Hebbian Plasticity in Random Networks}.
\newblock \emph{arXiv}, Jul 2020.
\newblock URL \url{https://arxiv.org/abs/2007.02686v4}.

\bibitem[Neumann et~al.(1966)Neumann, Burks, et~al.]{neumann1966theory}
J{\'a}nos Neumann, Arthur~W Burks, et~al.
\newblock \emph{Theory of self-reproducing automata}, volume 1102024.
\newblock University of Illinois press Urbana, 1966.

\bibitem[Nichele et~al.(2017)Nichele, Ose, Risi, and Tufte]{nichele2017neat}
Stefano Nichele, Mathias~Berild Ose, Sebastian Risi, and Gunnar Tufte.
\newblock Ca-neat: evolved compositional pattern producing networks for
  cellular automata morphogenesis and replication.
\newblock \emph{IEEE Transactions on Cognitive and Developmental Systems},
  10\penalty0 (3):\penalty0 687--700, 2017.

\bibitem[Pedersen \& Risi(2021)Pedersen and Risi]{pedersen2021evolving}
Joachim~Winther Pedersen and Sebastian Risi.
\newblock Evolving and merging hebbian learning rules: increasing
  generalization by decreasing the number of rules.
\newblock In \emph{Proceedings of the Genetic and Evolutionary Computation
  Conference}, pp.\  892--900, 2021.

\bibitem[Pugh et~al.(2016)Pugh, Soros, and Stanley]{pugh2016quality}
Justin~K Pugh, Lisa~B Soros, and Kenneth~O Stanley.
\newblock Quality diversity: A new frontier for evolutionary computation.
\newblock \emph{Frontiers in Robotics and AI}, 3:\penalty0 40, 2016.

\bibitem[Risi(2021)]{risi2021selfassemblingAI}
Sebastian Risi.
\newblock The future of artificial intelligence is self-organizing and
  self-assembling.
\newblock \emph{sebastianrisi.com}, 2021.
\newblock URL \url{https://sebastianrisi.com/self_assembling_ai}.

\bibitem[Risi \& Stanley(2013)Risi and Stanley]{risi2013confronting}
Sebastian Risi and Kenneth~O Stanley.
\newblock Confronting the challenge of learning a flexible neural controller
  for a diversity of morphologies.
\newblock In \emph{Proceedings of the 15th annual conference on Genetic and
  evolutionary computation}, pp.\  255--262, 2013.

\bibitem[Risi \& Stanley(2019)Risi and Stanley]{risi2019deep}
Sebastian Risi and Kenneth~O Stanley.
\newblock Deep neuroevolution of recurrent and discrete world models.
\newblock In \emph{Proceedings of the Genetic and Evolutionary Computation
  Conference}, pp.\  456--462, 2019.

\bibitem[Ruiz et~al.(2020)Ruiz, Vilalta, and Moreno-Noguer]{Ruiz2020Jun}
Alejandro~Hernandez Ruiz, Armand Vilalta, and Francesc Moreno-Noguer.
\newblock {Neural Cellular Automata Manifold}.
\newblock \emph{arXiv}, Jun 2020.
\newblock URL \url{https://arxiv.org/abs/2006.12155v3}.

\bibitem[Salimans et~al.(2017)Salimans, Ho, Chen, Sidor, and
  Sutskever]{Salimans2017Mar}
Tim Salimans, Jonathan Ho, Xi~Chen, Szymon Sidor, and Ilya Sutskever.
\newblock {Evolution Strategies as a Scalable Alternative to Reinforcement
  Learning}.
\newblock \emph{arXiv}, Mar 2017.
\newblock URL \url{https://arxiv.org/abs/1703.03864v2}.

\bibitem[Stanley \& Miikkulainen(2003)Stanley and
  Miikkulainen]{stanley2003taxonomy}
Kenneth~O Stanley and Risto Miikkulainen.
\newblock A taxonomy for artificial embryogeny.
\newblock \emph{Artificial life}, 9\penalty0 (2):\penalty0 93--130, 2003.

\bibitem[Stanley et~al.(2009)Stanley, D'Ambrosio, and
  Gauci]{stanley2009hypercube}
Kenneth~O Stanley, David~B D'Ambrosio, and Jason Gauci.
\newblock A hypercube-based encoding for evolving large-scale neural networks.
\newblock \emph{Artificial life}, 15\penalty0 (2):\penalty0 185--212, 2009.

\bibitem[Stanley et~al.(2019)Stanley, Clune, Lehman, and
  Miikkulainen]{stanley2019designing}
Kenneth~O Stanley, Jeff Clune, Joel Lehman, and Risto Miikkulainen.
\newblock Designing neural networks through neuroevolution.
\newblock \emph{Nature Machine Intelligence}, 1\penalty0 (1):\penalty0 24--35,
  2019.

\bibitem[Such et~al.(2017)Such, Madhavan, Conti, Lehman, Stanley, and
  Clune]{such2017deep}
Felipe~Petroski Such, Vashisht Madhavan, Edoardo Conti, Joel Lehman, Kenneth~O
  Stanley, and Jeff Clune.
\newblock Deep neuroevolution: Genetic algorithms are a competitive alternative
  for training deep neural networks for reinforcement learning.
\newblock \emph{arXiv preprint arXiv:1712.06567}, 2017.

\bibitem[Sudhakaran et~al.(2021)Sudhakaran, Grbic, Li, Katona, Najarro,
  Glanois, and Risi]{Sudhakaran2021Mar}
Shyam Sudhakaran, Djordje Grbic, Siyan Li, Adam Katona, Elias Najarro, Claire
  Glanois, and Sebastian Risi.
\newblock {Growing 3D Artefacts and Functional Machines with Neural Cellular
  Automata}.
\newblock \emph{arXiv}, Mar 2021.
\newblock URL \url{https://arxiv.org/abs/2103.08737v2}.

\bibitem[Tavares et~al.(2006)Tavares, Kreutzer, and
  Fedor]{Tavares2006NeuroCellularAC}
Jorge Tavares, Cornelia Kreutzer, and Anna Fedor.
\newblock Neuro-cellular automata: Connecting cellular automata, neural
  networks and evolution.
\newblock 2006.

\bibitem[Wolfram(2002)]{Wolfram2002}
Stephen Wolfram.
\newblock \emph{{A New Kind of Science}}.
\newblock Wolfram Media, 2002.
\newblock ISBN 978-1-57955008-0.

\bibitem[Wulff \& Hertz(1992)Wulff and Hertz]{wulff1992learning}
N~Wulff and J~A Hertz.
\newblock Learning cellular automaton dynamics with neural networks.
\newblock \emph{Advances in Neural Information Processing Systems}, 5:\penalty0
  631--638, 1992.

\bibitem[Zador(2019)]{zador2019critique}
Anthony~M Zador.
\newblock A critique of pure learning and what artificial neural networks can
  learn from animal brains.
\newblock \emph{Nature communications}, 10\penalty0 (1):\penalty0 1--7, 2019.

\bibitem[Zhmoginov et~al.(2022)Zhmoginov, Sandler, and
  Vladymyrov]{zhmoginov2022hypertransformer}
Andrey Zhmoginov, Mark Sandler, and Max Vladymyrov.
\newblock Hypertransformer: Model generation for supervised and semi-supervised
  few-shot learning, 2022.

\bibitem[Zwirn \& Delahaye(2011)Zwirn and Delahaye]{Zwirn2011Nov}
Herve Zwirn and Jean-Paul Delahaye.
\newblock {Unpredictability and Computational Irreducibility}.
\newblock \emph{arXiv}, Nov 2011.
\newblock URL \url{https://arxiv.org/abs/1111.4121v2}.

\end{thebibliography}
